\documentclass{article}
\usepackage{ijcai25}

\usepackage{times}
\usepackage{soul}
\usepackage{url}
\usepackage[hidelinks]{hyperref}
\usepackage[utf8]{inputenc}
\usepackage[small]{caption}
\usepackage{graphicx}
\usepackage{amsmath}
\usepackage{amsthm}
\usepackage{booktabs}
\usepackage{algorithm}
\usepackage{algorithmic}
\usepackage[switch]{lineno}
\usepackage{subcaption}

\title{IJCAI--25 Formatting Instructions}
\title{Exploring Multimodal Foundation AI and Expert-in-the-Loop for Sustainable Management of Wild Salmon Fisheries in Indigenous Rivers}

\author{
Chi Xu$^1$\and
Yili Jin$^{1,2}$\and
Sami Ma$^1$\and
Rongsheng Qian$^1$\and
Hao Fang$^1$\and
Jiangchuan Liu$^1$\and\\
Xue Liu$^2$\and
Edith C.H. Ngai$^3$\and
William I. Atlas$^4$\and
Katrina M. Connors$^5$\and
Mark A. Spoljaric$^6$\\
\affiliations
$^1$Simon Fraser University, Vancouver, Canada~~~$^2$McGill University, Montréal, Canada\\
$^3$The University of Hong Kong, Hong Kong, China~~~$^4$Wild Salmon Center, Portland, USA \\
$^5$Pacific Salmon Foundation, Vancouver, Canada~~~$^6$Haida Fisheries Program, Skidegate, Canada\\
\vspace{0.2cm}
\emails
chix@sfu.ca, yili.jin@mail.mcgill.ca, \{masamim, rqa4, fanghaof, jcliu\}@sfu.ca, \\ xueliu@cs.mcgill.ca, chngai@eee.hku.hk, watlas@wildsalmoncenter.org, \\ 
\vspace{0.05cm}
kconnors@psf.ca, mark.spoljaric@haidanation.com
}

\begin{document}

\maketitle

\begin{abstract}
Wild salmon are essential to the ecological, economic, and cultural sustainability of the North Pacific Rim. Yet climate variability, habitat loss, and data limitations in remote ecosystems that lack basic infrastructure support pose significant challenges to effective fisheries management. This project explores the integration of multimodal foundation AI and expert-in-the-loop frameworks to enhance wild salmon monitoring and sustainable fisheries management in Indigenous rivers across Pacific Northwest. By leveraging video and sonar-based monitoring, we develop AI-powered tools for automated species identification, counting, and length measurement, reducing manual effort, expediting delivery of results, and improving decision-making accuracy. Expert validation and active learning frameworks ensure ecological relevance while reducing annotation burdens. To address unique technical and societal challenges, we bring together a cross-domain, interdisciplinary team of university researchers, fisheries biologists, Indigenous stewardship practitioners, government agencies, and conservation organizations. Through these collaborations, our research fosters ethical AI co-development, open data sharing, and culturally informed fisheries management.

\end{abstract}

\section{Problem Statement}

Wild salmon are integral to the social-ecological systems of the North Pacific Rim. For over 10,000 years, they have supported thriving fisheries~\cite{yoshiyama1999history,carothers2021indigenous}, sustained local economies, enriched cultures, and maintained ecological balance~\cite{walsh2020relationships,brown2021sociocultural}. Yet rapid environmental changes driven by climate variability are threatening the resilience of salmon ecosystems~\cite{waples2008evolutionary,di2016multi,frolicher2018emerging}. Across their range, wild salmon populations have declined significantly, with increasingly unpredictable returns~\cite{kilduff2015changing,dorner2018spatial}. These declines pose significant threats to their long-term sustainability and the communities, such as Indigenous people, that depend on them~\cite{atlas2021indigenous}.

Sustaining salmon fisheries is further complicated by mixed-stock fisheries, which indiscriminately harvest co-migrating populations~\cite{walters2008report,moore2021conservation}, and by the high costs and logistical challenges of monitoring in remote, roadless areas of the Pacific Northwest~\cite{price2017canada}. These challenges have created the need for adaptive AI models and systems that support in-season management and selective terminal fisheries targeting healthy populations~\cite{atlas2021indigenous}. Such AI models and systems can bolster ecosystem resilience and maintain productivity through cycles of salmon abundance, even amidst climate change~\cite{schindler2015prediction}.





\begin{figure*}[t]
  \centering
  \begin{subfigure}[b]{0.31\linewidth}
    \includegraphics[width=\linewidth]{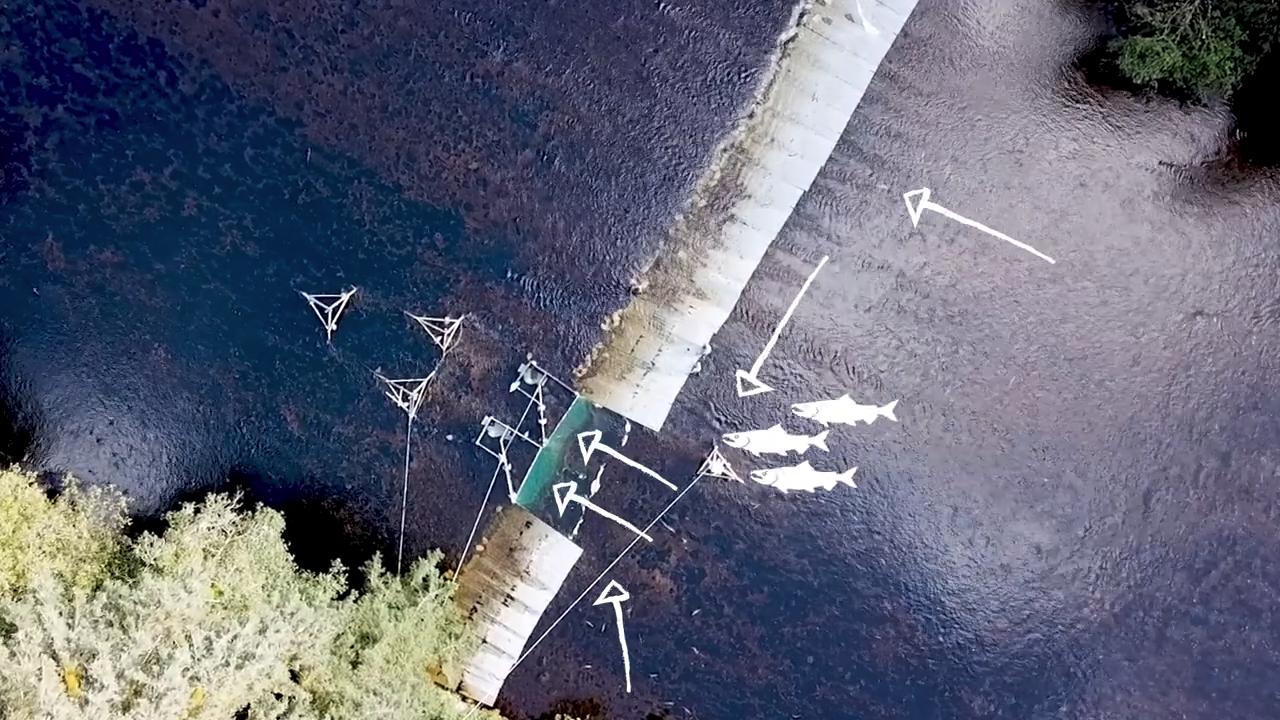}
    \caption{}
    \label{fig:weir-top}
  \end{subfigure}
  \hspace{0.01\linewidth}
  \begin{subfigure}[b]{0.30\linewidth}
    \includegraphics[width=\linewidth]{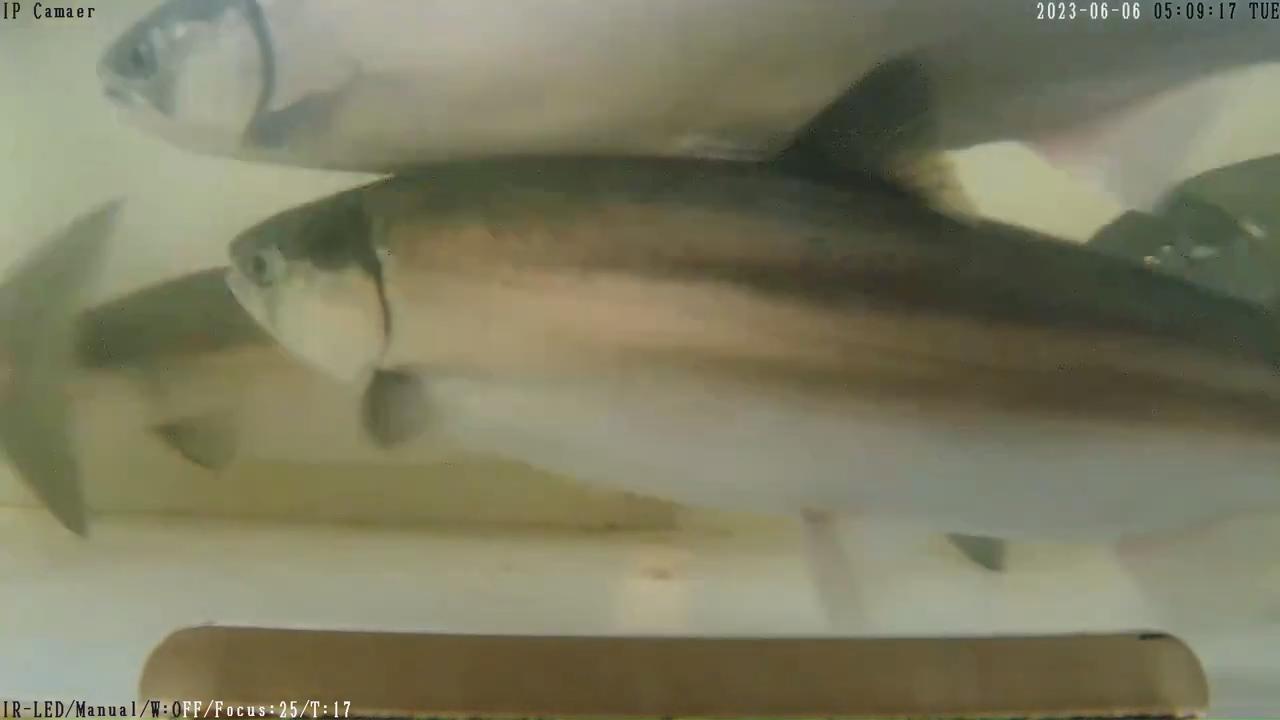}
    \caption{}
    \label{fig:sock}
  \end{subfigure}
      \hspace{0.01\linewidth}
    \begin{subfigure}[b]{0.34\linewidth}
    \includegraphics[width=\linewidth]{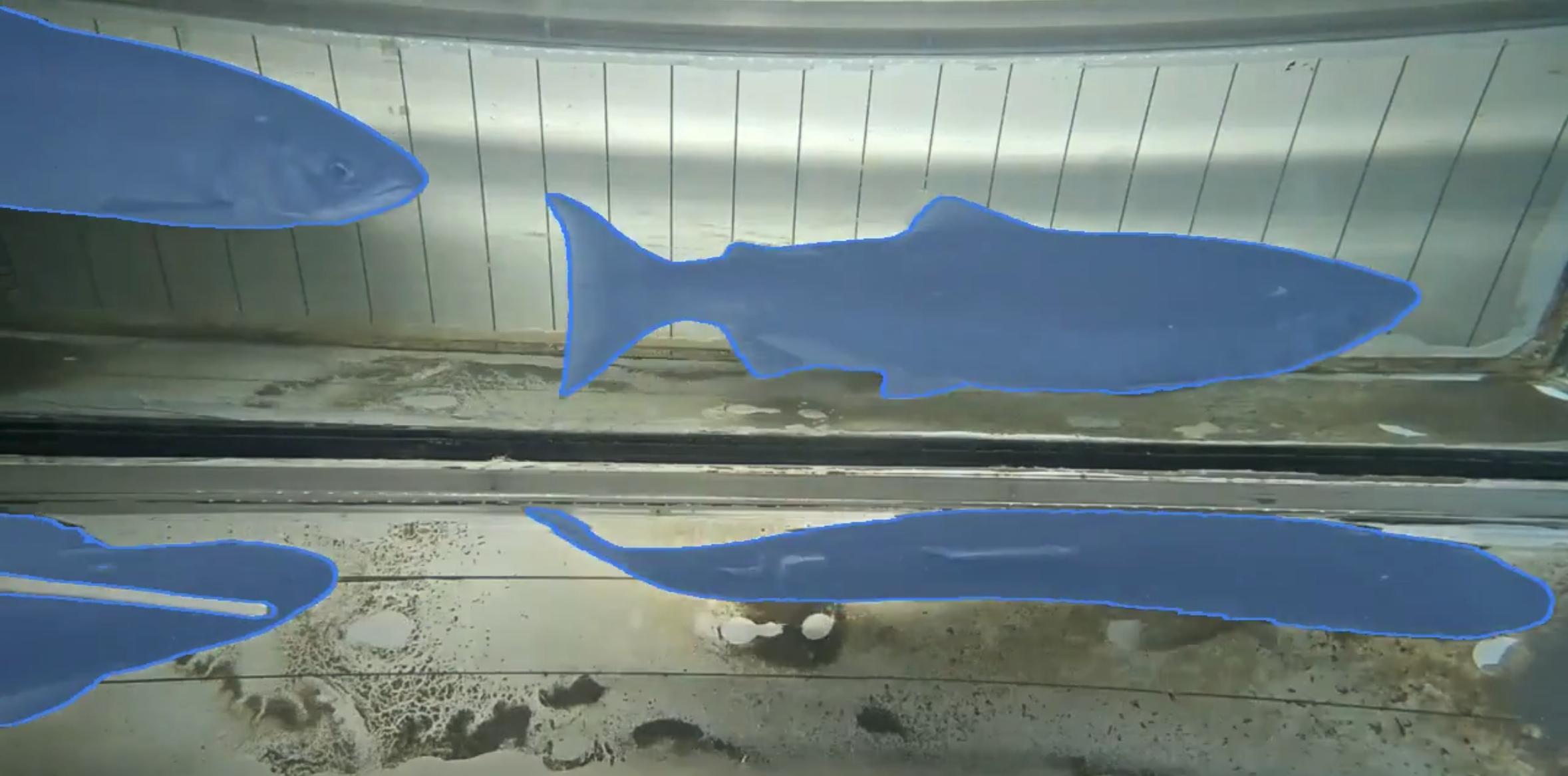}
    \caption{}
    \label{fig:segment}
  \end{subfigure}
    \vspace{-0.3cm}
  \captionsetup{width=0.9\linewidth}
  \caption{\textbf{(a)} A salmon counting weir at Koeye River (in Heiltsuk First Nation's traditional territory,  northern British Columbia) with salmon swimming passing the fish channel, \textbf{(b)} sample underwater video frames with salmon appearances, \textbf{(c)} object segmentation with species identification.}
  \label{fig:v1}
  \vspace{-0.3cm}
\end{figure*}

\begin{figure*}[t]
  \centering
    \begin{subfigure}[b]{0.20\linewidth}
    \includegraphics[width=\linewidth]{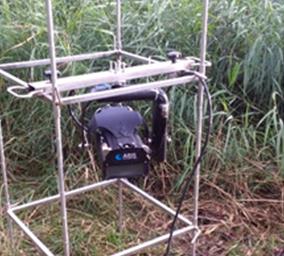}
    \caption{}
    \label{fig:sholder}
  \end{subfigure}
    \hspace{0.01\linewidth}
  \begin{subfigure}[b]{0.27\linewidth}
    \includegraphics[width=\linewidth]{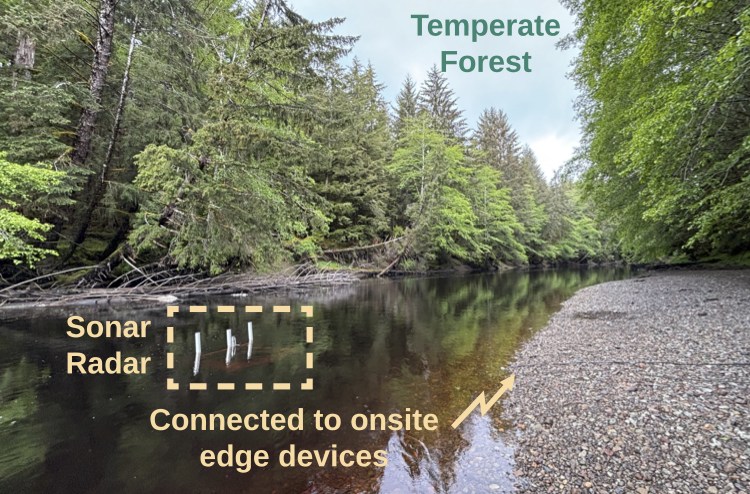}
    \caption{}
    \label{fig:haida}
  \end{subfigure}
  \hspace{0.01\linewidth}
  \begin{subfigure}[b]{0.27\linewidth}
    \includegraphics[width=\linewidth]{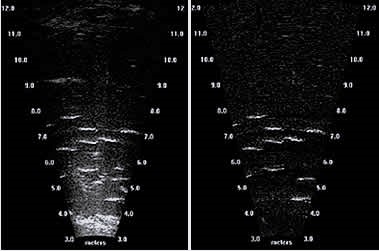}
    \caption{}
    \label{fig:sraw}
  \end{subfigure}
      \hspace{0.01\linewidth}
    \begin{subfigure}[b]{0.20\linewidth}
    \includegraphics[width=\linewidth]{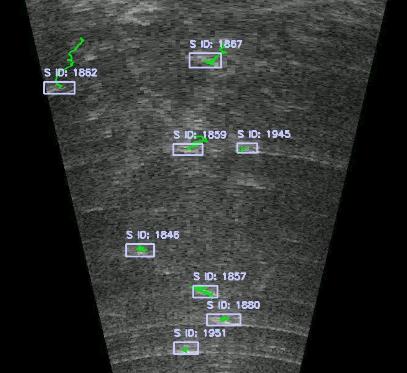}
    \caption{}
    \label{fig:strack}
  \end{subfigure}
    \vspace{-0.4cm}
  \captionsetup{width=0.9\linewidth}
  \caption{\textbf{(a)} A mounted ARIS sonar camera, \textbf{(b)} sonar deployment in the Yakoun River, Haida Nation's traditional territory, \textbf{(c)} sample frames from ARIS sonar, \textbf{(d)} salmon detection and tracking in sonar frames.}
  \label{fig:s1}
  \vspace{-0.3cm}
\end{figure*}

The initial effort to integrate computer vision and artificial intelligence into salmon monitoring focused on video-based weir systems aimed to expedite in-season fish counting~\cite{atlas2023wild}, a process traditionally requiring extensive manual effort. Early approaches leveraged underwater RGB cameras; however, the limitations of visual clarity and environmental variability necessitated alternative sensing modalities. Sonar-based monitoring emerged as a viable solution, exemplified by efforts such as Caltech’s CFC dataset~\cite{kay2022caltech}, which introduced fish detection, tracking, and counting in sonar videos. Unlike conventional \emph{Multi-Object Tracking} (MOT) datasets focused on urban environments, CFC highlights the challenges of domain generalization in low signal-to-noise underwater settings~\cite{kay2024align}. SALINA~\cite{xu2024salina} further extended these efforts by enabling real-time sonar analytics through Transformer-based models and energy-efficient deployment, supporting sustainable fisheries management in remote ecosystems within  Indigenous territories.

Despite these advancements, new challenges and opportunities remain. Moving beyond basic fish object detection, there is a growing need to integrate fishery management models and forecasting with motion-based counting, length measurement, and AI-assisted decision-making. Incorporating the newly developed vision foundation model~\cite{achiam2023gpt,liu2024visual} also helps improve performance and generate timely and accurate insights. 

For effective AI system deployment, ensuring reliable data capture, sufficient energy availability, and stable long-term operation is essential. However, the lack of basic infrastructure in the remote forests of the Pacific Northwest makes deployments particularly challenging~\cite{ma2024leo}. Therefore, integrating the expertise of Indigenous stewardship practitioners and fisheries biologists becomes even more critical to enhancing the system’s resilience and applicability. Furthermore, expert-in-the-loop frameworks and cross-referencing across multiple sensing modalities hold promise for improving accuracy, robustness, and actionable insights for sustainable fisheries management~\cite{wu2022survey}.

In response to this need, we created an interdisciplinary collaborative team to apply multimodal foundation AI and co-develop expert-in-the-loop frameworks for: (1) automated species identification and counting of salmon from video generated at salmon counting weirs, and (2) automated tracking, counting and length measurement from in-river sonar camera units. As shown in Figures~\ref{fig:v1} and ~\ref{fig:s1}, these two domains are complementary: video-based analysis can provide accurate estimates of species abundance passing through a fixed location, while sonar-based techniques can monitor salmon across the entire river width without the need to build special-purpose weirs and fish channels. 

A key objective of the project is to foster collaboration between university researchers, conservation practitioners, and Indigenous communities leading the stewardship of wild salmon populations within their territories. Multimodal data insights can improve our collective ability to detect, recognize and analyze videos of salmon and improve the understanding of the behavior and activities of different salmon species, for example run timing, abundance, and interannual variation in both. These insights will be instrumental towards the broader goal of preserving and conserving salmon and to understand the complex and interrelated factors that impact the health of these important species. Furthermore, data-driven fish passage improvements and habitat restoration can address biodiversity loss and ecosystem degradation.

\section{Alignment with Sustainable Development Goals and the LNOB principle}
The 2030 Agenda for Sustainable Development outlines 17 goals to promote global prosperity, equality, and sustainability. A core principle, Leaving No One Behind (LNOB), upholds the commitment that all communities, especially vulnerable groups, benefit from development efforts. Wildlife conservation plays a crucial role in sustainability, affecting both ecosystems and human livelihoods~\cite{liu24fish,gordon23rhinos,kshitiz23bird}. This research, emphasizing wild salmon monitoring, management, and conservation, aligns with several \emph{Sustainable Development Goals} (SDGs) as follows.

\begin{figure*}[t]
\centering
  \includegraphics[width=0.85\textwidth]{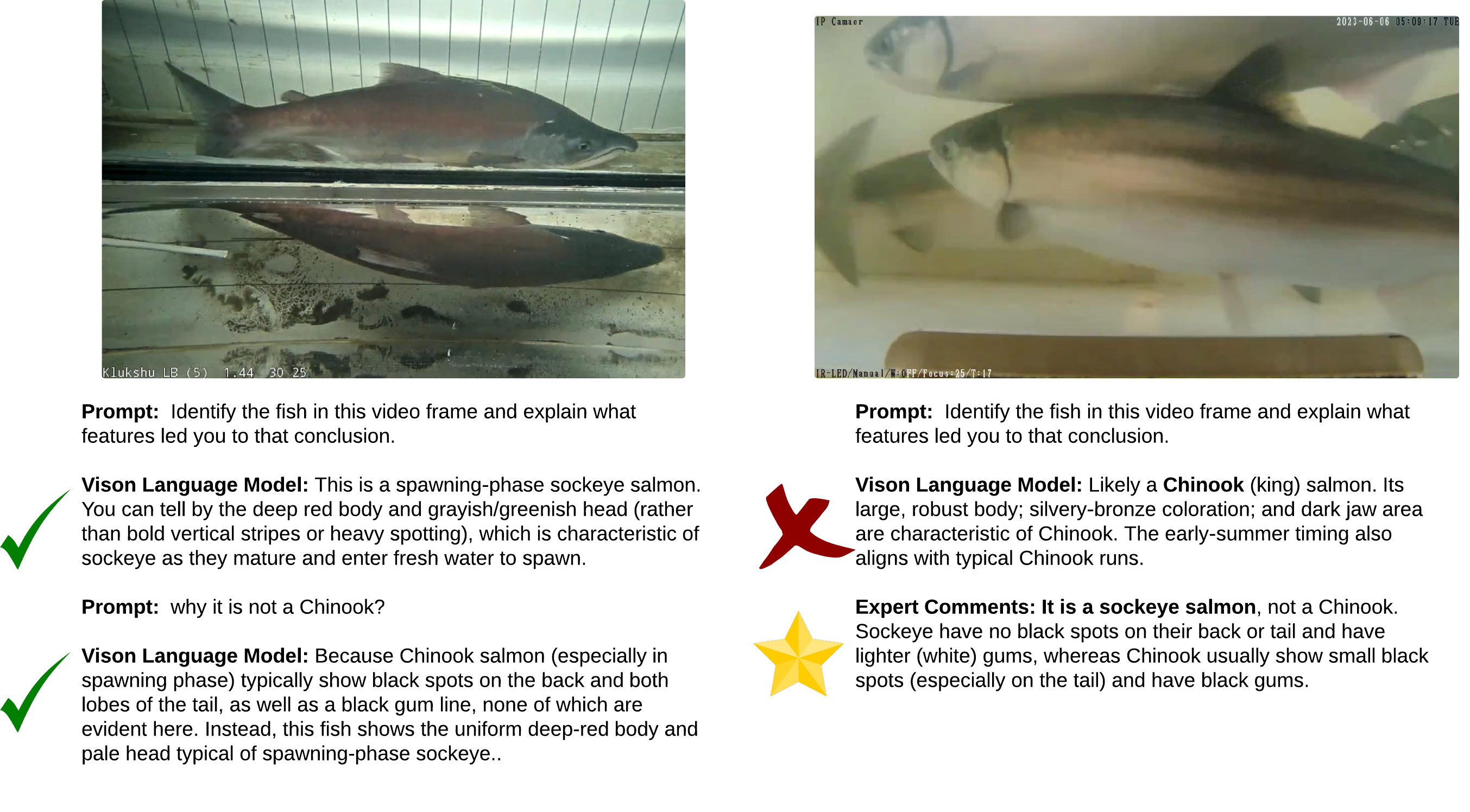}
    \vspace{-0.5cm}
\captionsetup{width=0.8\textwidth}
  \caption{An off-the-shelf vision language model (e.g., OpenAI o1) can identify some cases but also makes errors, thus requiring further refinement.}
  \label{fig:vlm}
  \vspace{-0.5cm}
\end{figure*}

\textbf{SDG 14: Life below Water.}
The multimodal foundation AI for monitoring and conserving wild salmon aligns with SDG 14.2 (Protect and restore ecosystems) by supporting habitat preservation and ecosystem resilience. Meanwhile, it contributes to SDG 14.4 (Regulate harvesting and end overfishing) by enabling sustainable fisheries management and aiding in the conservation of marine biodiversity.

\textbf{SDG 15: Life on Land.}
Wild salmon sustain both aquatic and terrestrial ecosystems. Their migrations transfer nutrients to forests and wildlife. Declining populations disrupt these cycles, affecting species such as bears and eagles. The management and conservation efforts in this project, such as data-driven fish passage improvements and habitat restoration, address biodiversity loss and degradation of terrestrial and inland freshwater ecosystem (aligning with SDGs 15.1 and 15.4).


\textbf{SDG 17: Partnerships for the Goals.}
Sustainable salmon conservation requires collaboration with government agencies, Indigenous rights holders, and diverse stakeholders. In this project, our interdisciplinary team include Indigenous communities, university researchers, conservation organizations, and industry partners. Integrating Indigenous knowledge with modern technologies, such as AI-powered monitoring, enhances conservation strategies and culturally informed approaches. These efforts align with SDG 17.16 (Strengthen partnerships through knowledge and resource sharing) and SDG 17.17 (Encourage multi-stakeholder collaborations).

\textbf{Alignment with LNOB.}
Wild salmon are essential to Indigenous culture, economy, and food security, yet declining stocks exacerbate food insecurity and economic hardship. LNOB promotes equitable conservation efforts by recognizing Indigenous communities as key decision-makers in managing local salmon populations, strengthening co-governance and long-term fishery access. AI-powered monitoring enhances data collection and fisheries management, contributing to ecosystem conservation. When integrated with traditional Indigenous knowledge, AI supports more culturally informed policies, sustainability efforts, and Indigenous sovereignty in conservation.

\section{Strategies and Methods}

\subsection{Multimodal Foundation AI for Salmon Monitoring}
Wild salmon monitoring requires accurate species identification, counting, and length measurement. Video-based and sonar-based approaches offer complementary advantages, yet each faces unique challenges. In this project, we explore multimodal foundation AI to improve model accuracy, reduce annotation burden, and enhance AI explainability.

Video-based monitoring at salmon counting weirs uses underwater RGB cameras to capture detailed visual features, yet occlusion, environmental variability, and data imbalance affect accuracy~\cite{khan2023fishnet}. Salmon frequently overlap in dense aggregations, making single-camera detection unreliable. To address this, we leverage multi-view fusion, where synchronized cameras or optical mirrors at different angles provide complementary perspectives. This approach reveals occluded fish objects and trajectories. Species identification also suffers from data skew, particularly when rare species are underrepresented. We apply data augmentation techniques such as synthetic image generation and class-balanced sampling~\cite{cui2019class} to mitigate dataset imbalance.

For low-confidence salmon detections and classifications in videos, we leverage pre-trained vision language models (e.g., LLaVA~\cite{liu2024visual}, gpt-4o~\cite{zhong2024evaluation}, and OpenAI o1 \cite{openai2024reason}) that generates natural language descriptions and species identifications. As shown in Figure~\ref{fig:vlm}, while an off-the-shelf vision language model can identify some cases, it also produces errors compared to expert feedback from fisheries biologists. Therefore, expert validation is necessary to refine model predictions further. Another issue we identified during annotation for this domain-specific task is that inexperienced annotators further introduce labelling errors, thus reducing model reliability. To enhance annotation quality and model interpretability, we also integrate vision language models into the annotation phase, helping to minimize errors.

Sonar-based monitoring in rivers enables salmon detection, tracking, and counting in turbid environments, but presents challenges in noise reduction, spatial-temporal modeling, and cross-modal integration~\cite{xu2024salina}. Sonar data contains substantial background noise due to environmental factors such as water turbulence and reflections. In this project, we employ deep learning-based denoising models~\cite{garber2024image,chihaoui2024masked} trained on sonar datasets to enhance signal clarity. Existing sonar-based tracking systems struggle with false positives and temporal inconsistencies. To improve salmon tracking performance across frames, we incorporate spatial-temporal features using transformer-based architectures, which will be further introduced in our implementation plan (Section 4.3). Notably, sonar frames alone lack species-level resolution, limiting classification accuracy. To address this, we synchronize sonar and video data when they are both available, integrating features through early fusion techniques. By aligning spatial and temporal cues with multimodal inputs, our approach enhances tracking and counting performance.


\begin{figure}
\centering 
  \includegraphics[width=0.42\textwidth]{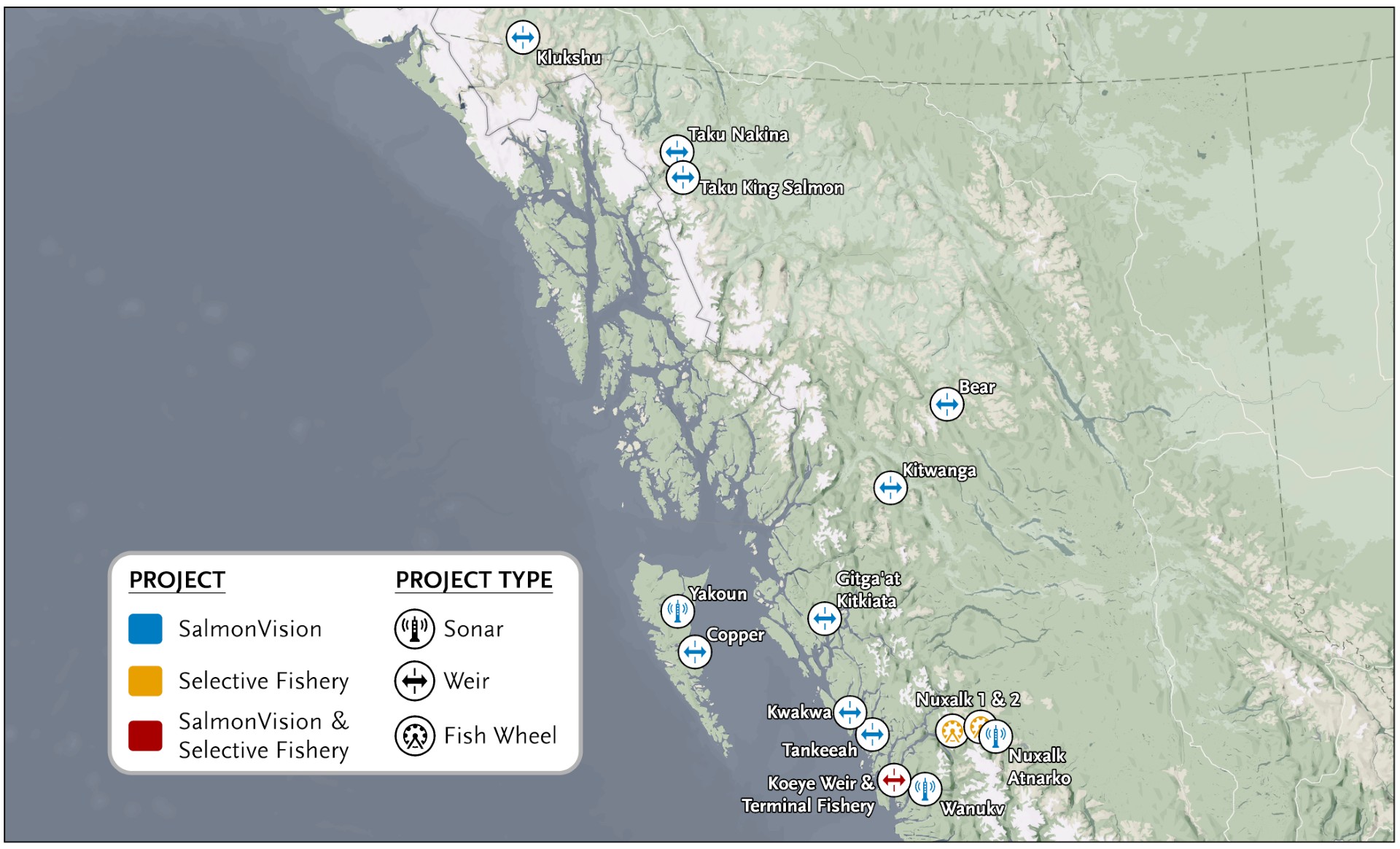}
  \caption{This project supports SalmonVision \& Selective Fishery in multiple Indigenous rivers in British Columbia, Canada.}
  \label{fig:bc}
  \vspace{-0.3cm}
\end{figure}

\subsection{Scalable and User-Centric Framework Design}
Scalability and real-time performance are critical for deploying AI-powered monitoring across different Indigenous rivers to generate in-season insights for fisheries management.  Both video and sonar data streams generate large volumes of high-dimensional data, necessitating efficient processing pipelines. Limited power, computation resources, and network connectivity in the remote forests of the Pacific Northwest further exacerbate challenges~\cite{xu2024salina}. To this end, we have implemented a hybrid edge-cloud architecture where lightweight models perform initial detection on edge devices, reducing computational demand and transmission costs. More complex tasks, such as fine-grained species classification and anomaly detection, are offloaded to cloud servers. Such task offloading maintains computational efficiency without sacrificing accuracy. To further improve model adaptability, we incorporate continual learning mechanisms that update model parameters based on multi-year data while preserving previously learned patterns.

In summary, our framework design enables on-site AI inference while maintaining remote access for diverse stakeholders. Edge-based computing allows immediate fish detection and tracking at monitoring sites, even in locations with limited internet connectivity. Meanwhile, cloud integration supports large-scale data storage, remote model updates, and collaborative access to processed data. We also explore federated learning~\cite{liu2020fedvision}, which enhances privacy and Indigenous data sovereignty by enabling model improvements without direct data transfer between monitoring sites.


The adoption of AI-powered monitoring tools also depends on usability. We design a user-friendly application interface that enables Indigenous fisheries biologists to access real-time AI-generated insights with minimal technical expertise. Standardized hardware and software integration supports seamless deployment across different monitoring locations within Indigenous territories, as shown in Figure~\ref{fig:bc}. To further promote accessibility, we provide open-source documentation and training resources, allowing fisheries practitioners to deploy and maintain the system without specialized AI knowledge. By integrating scalable computing solutions with intuitive design, we create an adaptive and inclusive monitoring framework that incorporates fisheries experts in the loop.


\subsection{Collaboration and Open Data for Effective Fisheries Management}
Effective fisheries management requires transparent data sharing, cross-sector collaboration, and adaptive decision-making. We commit to open-sourcing datasets and models, encouraging innovation within the AI and fisheries research communities. By partnering with government agencies, Indigenous groups, and conservation organizations, we aim for monitoring strategies that are both scientifically rigorous and culturally informed. 

Integrating multimodal foundation AI, real-time monitoring systems, and expert-in-the-loop frameworks, we transition fisheries management from a paradigm defined by data limitations or dependent on unreliable preseason forecasts to adaptive in-season decision-making. This shift enables fisheries experts and managers to respond dynamically to changing environmental conditions, improving conservation outcomes and increasing fishery opportunities when appropriate. Our findings will be disseminated through peer-reviewed conferences and journals in both AI and fisheries communities. Additionally, we will release open-access repositories\footnote{https://github.com/Salmon-Computer-Vision} to maximize accessibility and impact.  Through technological innovation and collaborative partnerships, we build a resilient, data-driven framework for sustainable management of wild salmon fisheries.


\section{Implementation Plan}

\subsection{Expert-in-the-Loop AI for Salmon Monitoring}

Building on our previous work, we have developed the SalmonVision web app\footnote{https://salmonvision.org/}, which enables user-led data review and annotation of salmon detection and classification results generated on the edge. This project extends that groundwork by exploring multimodal foundation AI to transform video and sonar data into actionable insights for salmon monitoring, also emphasizes the critical role of human expertise in the loop. We are currently developing SalmonVision web app to include the following features: 1) data collected from different monitoring sites within Indigenous territories is processed using advanced AI models capable of integrating multiple modalities, 2) AI-generated outputs (detections, counts, and species classifications) are further refined through experts' multimodal input, including dot annotations, bounding boxes, and text prompts.

The AI-human collaborative workflow ensures that expert knowledge informs every stage of the process. Fisheries experts contribute their domain expertise by validating and enhancing annotations and AI-generated outputs, creating a rich dataset of labeled frames for AI model refinement. This iterative process strengthens the AI model’s ability to generalize and perform reliably in real-world conditions, enabling it to better adapt to the unique environmental characteristics of each site within Indigenous territories. Once refined, the AI model is updated on edge-computing systems installed at the monitoring site. These systems operate autonomously to analyze incoming data in real time, but the process remains firmly anchored by human oversight. Fisheries practitioners provide ongoing feedback and technical support to support continuous system operation and data accuracy throughout the monitoring season.

\begin{figure}[t]
\centering
  \includegraphics[width=0.40\textwidth]{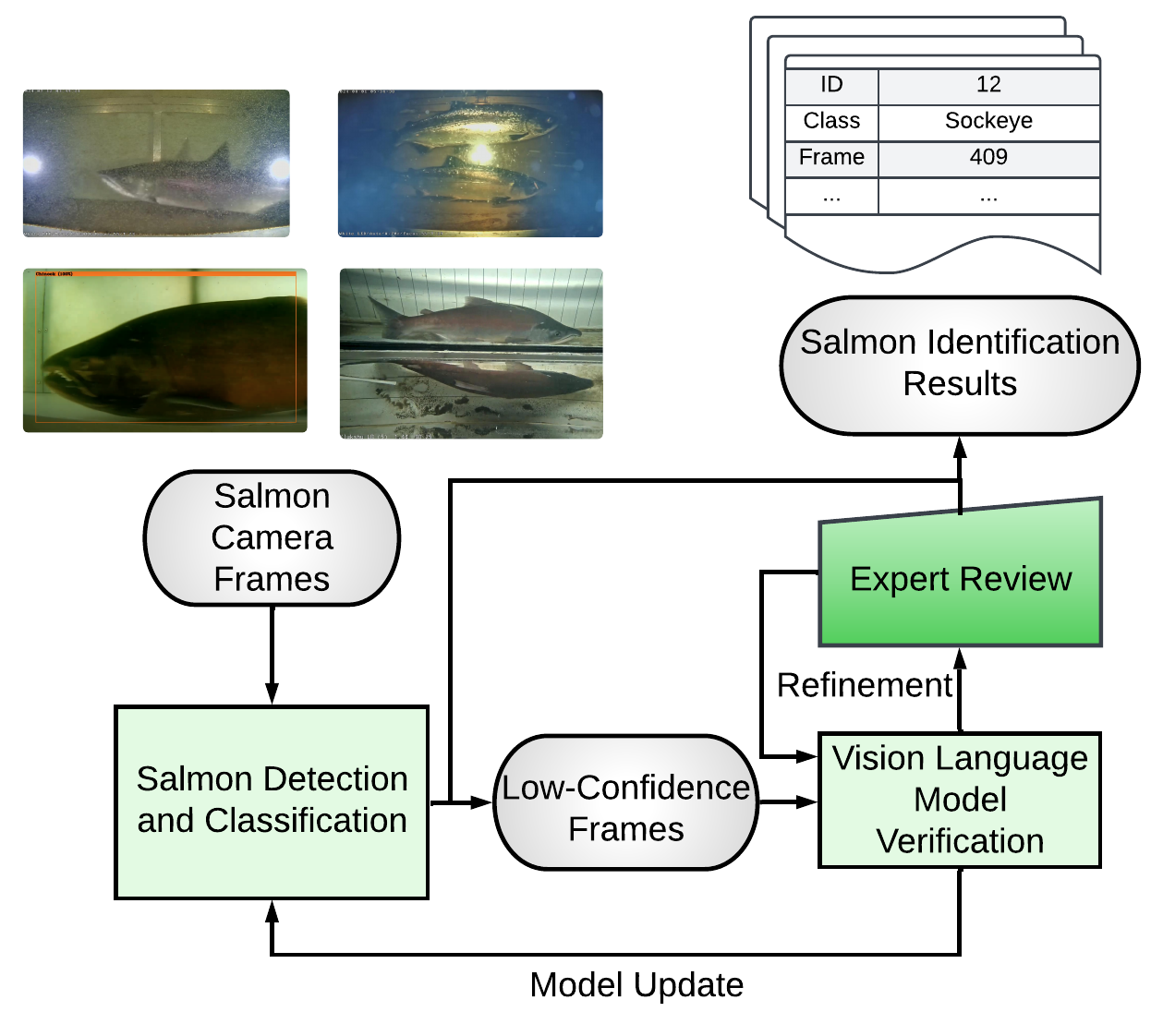}
    \vspace{-0.1cm}
  \caption{Vision language model verification and refinement.}
  \label{fig:vlmm}
  \vspace{-0.3cm}
\end{figure}
\subsection{Video-based Salmon Detection, Classification, and Counting}
The existing system at our salmon counting weir employs single-modality models such as YOLO~\cite{wang2024yolov10} and RT-DETR~\cite{zhao2024detrs} for salmon detection, classification, and counting. While these models achieve reasonable performance, they struggle with cases involving occlusions, poor lighting, or rare species due to their reliance on visual features alone. Misclassifications and low-confidence detections introduce errors that require extensive manual verification. To address these limitations, we incorporate a vision language model (VLM) that enhances explainability and integrates additional modalities, improving both detection and classification accuracy and expert review efficiency.

As shown in Figure~\ref{fig:vlmm}, our implementation leverages a VLM to refine low-confidence cases where the confidence scores of the base detection and classification model are low. Instead of relying solely on pixel-based features, the VLM generates descriptive textual explanations of its predictions, providing interpretable insights into classification decisions. This process may involve prompt engineering to guide the model in handling specific challenges such as distinguishing between visually similar species. When uncertainty remains high, both the text-based explanations and the corresponding video frames are flagged for expert review. By combining human expertise with model-driven reasoning, we facilitate the correction of misclassifications and their incorporation into the model’s continuous learning process.

As expert-reviewed frames accumulate, the refined VLM progressively improves its performance, reducing reliance on manual verification over time. The system transitions from a semi-automated workflow toward an AI-driven foundation model that can replace single-modality approaches. This shift enhances scalability by enabling high-accuracy fish monitoring with minimal human intervention. 

\begin{figure}
\centering
  \includegraphics[width=0.48\textwidth]{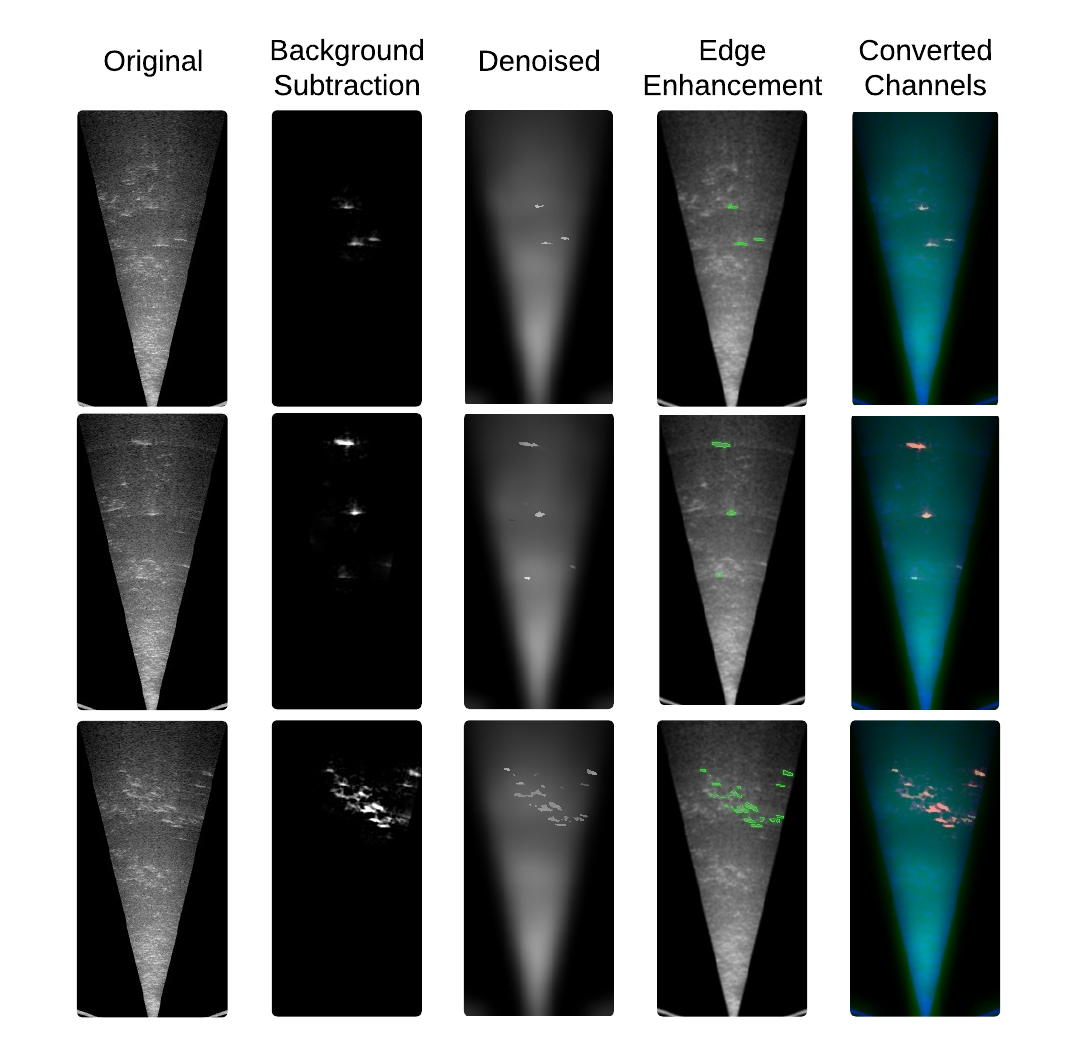}
  \caption{Traditional preprocessing of sonar frames.}
  \label{fig:sonar-pre}
  \vspace{-0.3cm}
\end{figure}

\subsection{Sonar-based Monitoring}

\begin{figure}
\centering 
  \includegraphics[width=0.4\textwidth]{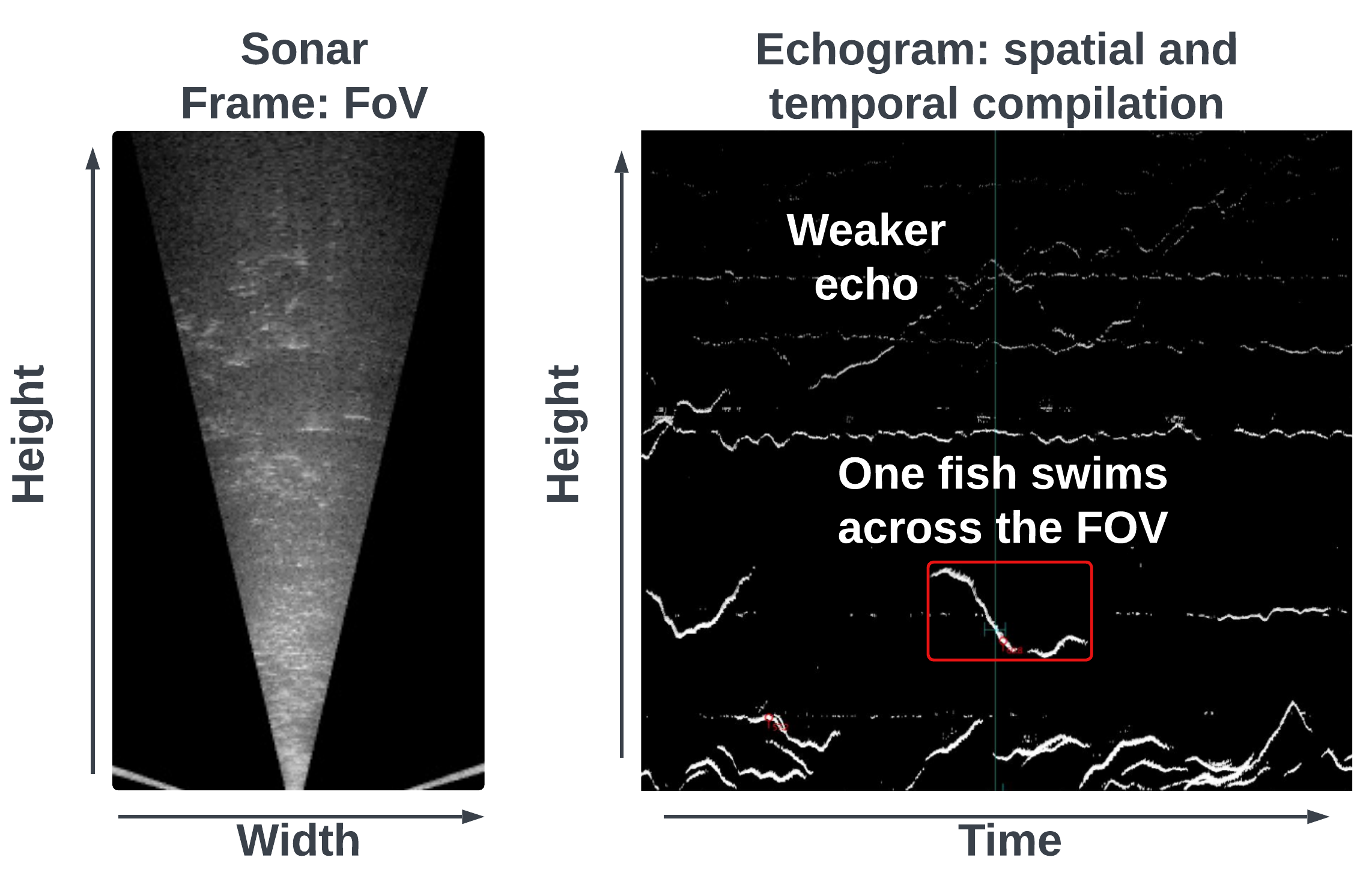}
  \caption{Sonar frames and echogram, as two different modalities.}
  \label{fig:echogram}
  \vspace{-0.3cm}
\end{figure}

The current sonar-based monitoring system in Indigenous rivers relies on traditional preprocessing techniques to enhance image quality for expert review and AI inference, as shown in Figure~\ref{fig:sonar-pre}. However, these preprocessing methods often introduce frame distortion and feature loss, degrading the performance of downstream tasks such as salmon detection, tracking, counting, and length measurement. To address these challenges, we propose adapting SAM2~\cite{ravi2024sam}, a recently published foundation model for these downstream tasks with multimodal inputs, including sonar frames and echograms. As shown in Figure~\ref{fig:echogram}, a sonar echogram is a time-series visualization of sonar returns, representing how acoustic signals interact with underwater objects and the riverbed over time.  The echogram serves as a key multimodal input, providing both spatial and temporal information to the adapted foundation model. By integrating echograms with sonar frames, we improve robustness against noise and enhance the performance of downstream tasks.

The first phase of our implementation focuses on integrating multimodal data to improve representation learning. Sonar frames and echograms are fused to create a more comprehensive input representation. To suppress noise while preserving critical information, we employ a lightweight foundation model such as CLIP~\cite{radford2021learning} to encode sonar frames and echograms into token representations. Unlike traditional denoising techniques, this approach reduces artifacts and prevents loss of essential details.

\begin{figure}
\centering 
  \includegraphics[width=0.48\textwidth]{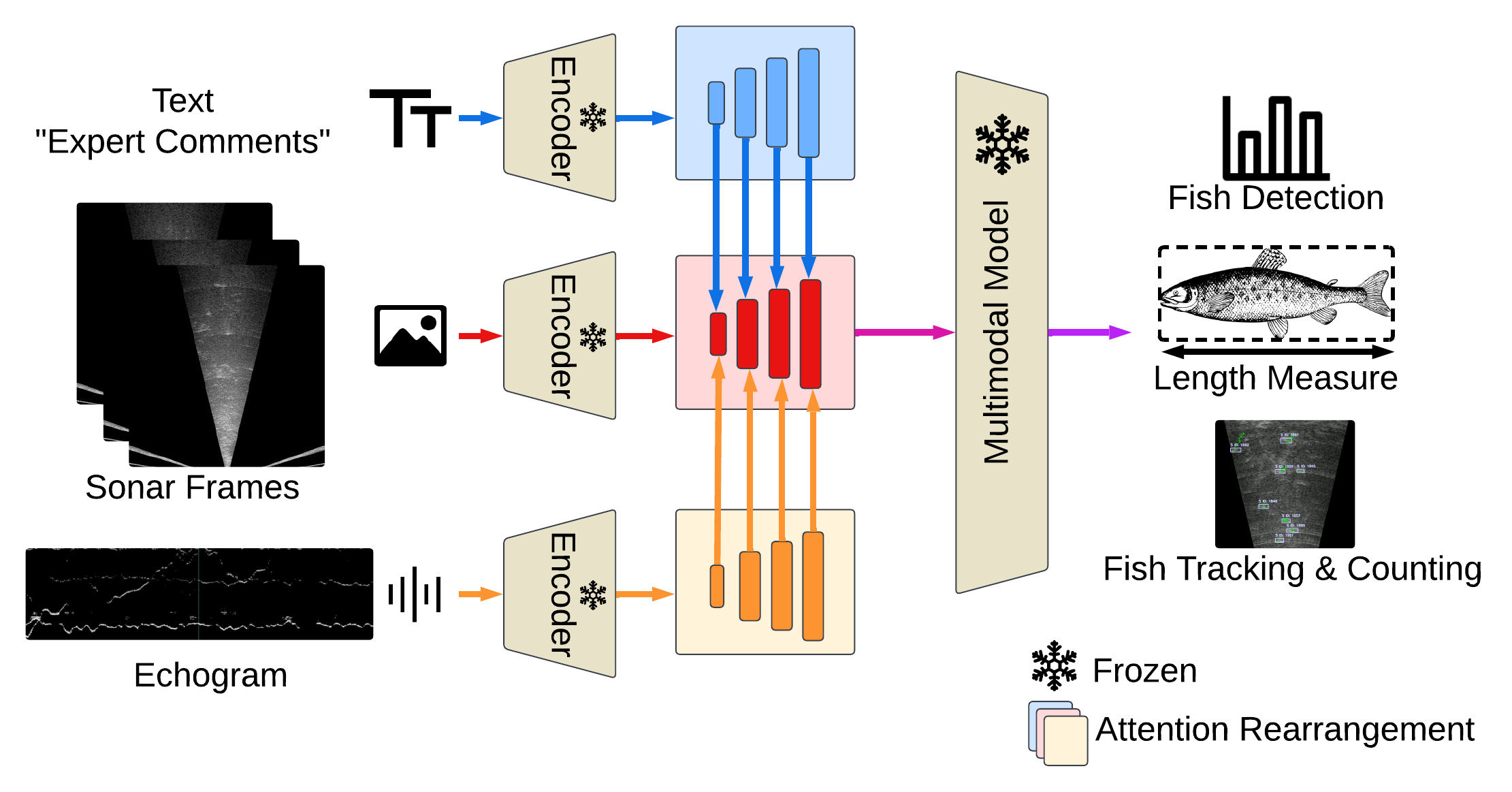}
  \captionsetup{width=0.9\linewidth}
  \caption{Multimodal foundation model for sonar domain.}
  \label{fig:framework}
  \vspace{-0.3cm}
\end{figure}

The second phase focuses on adapting the model to sonar-specific tasks and improving performance through fine-tuning. A labeled dataset of sonar frames and echograms, annotated with salmon appearances, positions, and length estimates, is prepared for training.

We apply transfer learning to the adapted SAM2, freezing early layers while fine-tuning later layers for sonar data. As shown in Figure \ref{fig:framework}, sonar frames and echograms are processed through separate encoders, facilitating effective multimodal integration. The extracted features are combined using attention-based fusion, incorporating expert comments as additional inputs when available. This structured fusion approach enhances the model’s ability to generalize under varying noise and environmental conditions, improving its adaptability to real-world sonar data.

For salmon detection, the adapted SAM2 generates segmentation masks or bounding boxes, refined using non-maximum suppression (NMS) to remove redundant predictions. Tracking and counting are performed using DeepSORT~\cite{wojke2017simple}, which combines motion and appearance features to ensure stable object association across frames. Centerline extraction is conducted using skeletonization algorithms, which refine fish contours for improved structural analysis. Length measurement is achieved through attention-based feature rearrangement, where extracted features are mapped to real-world metrics using known scaling factors. This approach ensures robust and accurate analysis of salmon populations in challenging underwater environments by leveraging both sonar frames and echogram signals.



\section{Evaluation Criteria}

\subsection{Video-Based Salmon Species Identification and Counting}
For species identification, we evaluate multi-class classification performance using mean Average Precision at IoU 0.5 (mAP@50) and F1 score~\cite{goodfellow2016deep}. These metrics assess precision and recall, with robustness across different species. We compare our VLM-enhanced identification results against baseline performance of single-modality models such as YOLO~\cite{wang2024yolov10} and RT-DETR~\cite{zhao2024detrs}, measuring improvements in classification performance. For fish counting, we assess the total counting accuracy using Mean Absolute Percentage Error (MAPE) and F1 score, which represents reliable enumeration in dense and occluded scenarios. Our approach is benchmarked against standard detection-based counting models, with improvements evaluated in terms of both precision and computational efficiency. We also compare VLM-enhanced species counts to expert-reviewed species counts to provide insight into VLM-enhanced salmon counting in real-world deployments.

\subsection{Sonar-Based Salmon Monitoring}
For sonar-based salmon detection performance, we also report mAP@50 and mAP@50:75, with the latter further capturing precision-recall trade-offs across different IoU thresholds. These metrics quantify the effectiveness of our multimodal approach in handling sonar-specific challenges such as substantial noise and low contrast. We compare against traditional contour-based methods and state-of-the-art baselines, including CFC-YOLO~\cite{kay2022caltech}, RT-DETR~\cite{zhao2024detrs}, and STSVT~\cite{xu2024salina}.

For tracking evaluation, we employ Multiple Object Tracking Accuracy (MOTA~\cite{bernardin2008evaluating}), Higher Order Tracking Accuracy (HOTA~\cite{luiten2021hota}), and IDF1~\cite{ristani2016performance} as benchmarks. MOTA quantifies the trade-off between missed detections, false positives, and identity switches, while HOTA incorporates temporal consistency and object association. IDF1 evaluates the accuracy of maintaining consistent object identities over time.
For fish counting, we evaluate numerical accuracy using Mean Average Error (MAE) and Root Mean Squared Error (RMSE), comparing the model with expert counts. For fish length estimation, we apply similar MAE and RMSE metrics to measure the deviation between model-predicted and manually measured fish lengths. Evaluations are also conducted against expert-reviewed sonar data to validate its reliability in real-world deployment. These evaluation criteria ensure rigorous assessment of our adapted multimodal foundation model’s performance, validating its effectiveness for real-time fisheries management, and allowing fine tuning for optimal performance at a specific site within Indigenous territories.

\section{Expected Results and Impacts}
Diverse foundation models with multimodal inputs are transforming society at an unprecedented rate; however, these AI models have rarely been co-developed with local or Indigenous communities. In this project, we co-develop AI models in collaboration with Indigenous communities, government agencies, and conservation practitioners across the North Pacific Rim. Our work aims to create lasting benefits for fisheries management and conservation while supporting equitable co-governance, empowering communities as decision-makers and stewards of local salmon populations.


Across the Pacific Northwest, thousands of locally adapted wild salmon populations remain unmonitored, despite being actively targeted in ongoing fisheries across marine and freshwater ecosystems. In an era of rapid climate change with no historical precedent, our work advances fisheries management by integrating multimodal foundation AI, real-time monitoring, and expert validation to enable adaptive, data-driven decision-making. Enhanced accuracy in fish population assessments will strengthen conservation strategies, establish management benchmarks for previously data-limited populations, and mitigate overfishing risks while supporting sustainable harvest opportunities.

Open-sourcing our datasets and models will accelerate research and innovation, fostering collaboration across AI, fisheries science, and conservation communities. Our cross-domain, interdisciplinary team will ensure that monitoring strategies are both scientifically rigorous and culturally relevant, enabling the translation of research into actionable fisheries management outcomes. By shifting from preseason forecasting to adaptive in-season management, this project will provide resilient, responsive tools for the sustainable management of wild salmon fisheries in an increasingly dynamic environment.

\section{Assumptions and Risks} 

Applying multimodal foundation AI to wild salmon monitoring and fisheries management is an emerging field, introducing inherent risks. In particular, the performance of such AI models for automated detections, tracking, counting, and length measurement across different sites remains uncertain, requiring multiple iterations of training and testing to meet fisheries management standards. However, our preliminary results from the Yakoun River suggest that automated model analysis is not only feasible but also critical for efficiently reviewing multimodal data. To mitigate these risks, we incorporate multiple rounds of data annotation, model training, testing, expert verification, and refinement to iteratively improve performance.


In addition, consideration of data ownership and AI-related risks for partnering with Indigenous communities is crucial to the ethical co-development of multimodal foundation AI. To address these concerns, we adopt an iterative co-development process, through which partner First Nations receive regular updates, provide input on research outcomes and tool development, and have their concerns addressed. Data-sharing agreements are established between our collaborative team and each Indigenous community, safeguarding their ownership of raw data outputs while permitting access to labeled data for model training and open research. Communities retain the right to withdraw from these agreements and remove their data from open-source repositories, though this has not been an issue to date.


\appendix

\section*{Project Team Description}

\noindent \textbf{Chi Xu} \\
\noindent \textit{Ph.D student, Simon Fraser University, Canada}\\
Chi Xu is currently pursuing his Ph.D. degree in computing science at Simon Fraser University, Canada. His research focuses on multimodal data sensing, management, and analytics, spanning the fields of AIoT and networked systems. \\

\noindent \textbf{Yili Jin} \\
\noindent \textit{Ph.D student, McGill University, Canada}\\
Yili Jin is a Ph.D. student in Computer Science at McGill University and is currently a visiting scholar at Simon Fraser University. His research interests include multimedia systems and communication, with a focus on applications for social good.\\

\noindent \textbf{Sami Ma} \\
\noindent \textit{Ph.D student, Simon Fraser University, Canada}\\
Sami Ma received the B.Sc. degree (Hons.) in computing science from Simon Fraser University, Burnaby, BC, Canada, in 2019, where he is currently pursuing his Ph.D. degree in computing science. His research interests include low earth orbit satellite networks, Internet architecture and protocols, deep learning, and computer vision.\\

\noindent \textbf{Rongsheng Qian} \\
\noindent \textit{Master student, Simon Fraser University, Canada}\\
Rongsheng Qian received a B.Sc. degree in Computing Science from Simon Fraser University, Burnaby, BC, Canada, in 2023. He is currently pursuing a Master's degree in Computing Science at the same institution. His research interests include deep learning and computer vision.\\

\noindent \textbf{Hao Fang} \\
\noindent \textit{Ph.D student, Simon Fraser University, Canada}\\
Hao Fang received the B.Sc. (Hons.) degree in computing science from Simon Fraser University, Burnaby, BC, Canada, in 2022, where he is currently pursuing the Ph.D. degree in computing science. His research interests include satellite communications and networking, particularly with multimedia systems. \\

\noindent \textbf{Dr. Jiangchuan Liu}\\
\noindent \textit{Professor, Simon Fraser University, Canada}\\
Dr. Jiangchuan Liu is currently a Professor at Simon Fraser University, Burnaby, BC, Canada. He is a Fellow of IEEE, and a Fellow of the Canadian Academy of Engineering. He has served on the editorial boards of IEEE/ACM Transactions on Networking, IEEE Transactions on Multimedia, IEEE Communications Surveys and Tutorials, and the IEEE Internet of Things Journal. He was a steering committee member of IEEE Transactions on Mobile Computing and a steering committee chair of IEEE/ACM IWQoS. He was the TPC Co-Chair of IEEE INFOCOM 2021 and the General Co-Chair of INFOCOM 2024.\\

\noindent \textbf{Dr. Xue Liu}\\
\noindent \textit{Professor, McGill University, Canada}\\
Dr. Xue Liu is a Full Professor and William Dawson Scholar at McGill University, with a courtesy appointment in the Department of Mathematics and Statistics. He is also the Associate Vice President of Research at Mohamed bin Zayed University of Artificial Intelligence (MBZUAI). In addition to his academic achievements, Professor Liu has held prominent industry roles, developing innovative research and technology and bridging them with impactful practical applications. Notably, from 2019 to 2024, he was Vice President of R\&D, Chief Scientist, and Co-Director at Samsung AI Center Montréal. From 2016 to 2019, he served as the Chief Scientist for Tinder Inc. Professor Liu is a Fellow of both the IEEE and the Canadian Academy of Engineering.\\ 

\noindent \textbf{Dr. Edith C. H. Ngai}\\
\noindent \textit{Associate Professor, The University of Hong Kong, China}\\
Dr. Ngai is currently an associate professor with the Department of
Electrical and Electronic Engineering, University of Hong Kong. Her research interests include Internet-of-Things, edge intelligence, smart cities, and smart health. She was a VINNMER fellow (2009) awarded by the Swedish Governmental Research Funding Agency VINNOVA. She was an area editor of the IEEE Internet of Things Journal from 2020 to 2022. She is currently an associate editor of IEEE Transactions of Mobile Computing, IEEE Transactions of Industrial Informatics, Ad Hoc Networks, and Computer Networks. She has served as a program chair in IEEE ISSNIP 2015, IEEE GreenCom 2022, and IEEE/ACM IWQoS 2024. She received a Meta Policy Research Award in Asia Pacific in 2022. She was selected as one of the N$^{2}$ Women Stars in computer networking and communications in 2022.\\

\noindent \textbf{Dr. William I Atlas}\\
\noindent \textit{Salmon Watershed Scientist, Wild Salmon Center, USA}\\
\noindent Dr. Atlas is a Salmon Watershed Scientist at the Wild Salmon Center (WSC). Prior to joining WSC in 2020, he worked with Central Coast First Nations (CCFN) for a decade, co-developing community-based salmon science as a graduate student and professional, and leading the development of the Central Coast Monitoring Framework as a postdoc with Pacific Salmon Foundation (PSF). In addition, he serves as the Central Coast Rep on the Northern Panel of the PSC and is an active member of the First Nations Caucus of the PSC, providing regular updates to community leadership and strategic support for CCFN engagement at bilateral management tables.\\

\noindent \textbf{Dr. Katrina Connors}\\
\noindent \textit{Director of Salmon Watersheds Program, Pacific Salmon Foundation (PSF), Canada}\\
Dr. Connors is the founding director of PSF’s Salmon Watersheds Program, a conservation science initiative working at the science-policy interface on Pacific salmon conservation and management issues. Katrina has over 19 years' experience leading collaborative research initiatives focused on improving our understanding of status and trends in Pacific salmon populations and cumulative pressures on their freshwater habitats. Katrina is also a Canadian Commissioner to the Pacific Salmon Commission (PSC). Katrina has overseen the successful delivery of several major collaborative grants including funding provided by the Coastal Restoration Fund and BCSRIF to expand the development of the Pacific Salmon Explorer to all salmon-bearing watersheds in BC.\\

\noindent \textbf{Mark A. Spoljaric}\\
\noindent \textit{Program Biologist, Haida Fisheries Program, Canada}\\
Mark A. Spoljaric is a fisheries biologist with over a decade of experience developing and managing salmon research and conservation initiatives on Haida Gwaii. As a Program Biologist at the Haida Fisheries Program, he oversees indicator stock assessments, escapement surveys, and project planning for multiple salmon-bearing watersheds. Mark has also conducted habitat assessments in remote coastal regions, enumerated returning salmon and juvenile fish, and contributed to public outreach on sustainable aquaculture practices.

\section*{Ethical Statement}
This research is committed to ethical, inclusive, and culturally respectful practices. We actively collaborate with Indigenous groups to ensure that fisheries monitoring strategies integrate traditional ecological knowledge and respect Indigenous data sovereignty. All data collection and analysis will follow established ethical guidelines, with consent-driven participation and transparency in data use. 
We prioritize fair representation, ensuring that stakeholders—including Indigenous communities, government agencies, and conservation organizations—are equitably involved in decision-making and benefit from the research outcomes. 

Additionally, we adhere to responsible AI principles, ensuring that machine learning models are interpretable, unbiased, and aligned with conservation goals. By fostering open collaboration and ethical data practices, this research contributes to sustainable fisheries management while upholding scientific integrity and social responsibility.

\section*{Acknowledgments}

This research is supported by an NSERC Discovery Grant, a British Columbia Salmon Recovery and Innovation Fund (BCSRIF\_2022\_401), and a MITACS Accelerate Cluster Grant, and received additional funding support from experiment.com. 
We are grateful for the trust and collaboration of the Heiltsuk, the Haida, Kitasoo Xai'xais, Taku River Tlingit and Gitga'at First Nations, as well as the Skeena Fishery Commission and Gitanyow Fisheries Authority for their ongoing partnership in this work.

\bibliographystyle{named}
\bibliography{ijcai25}

\end{document}